\documentclass[lettersize,journal]{IEEEtran}
\usepackage{amsmath,amsfonts}
\usepackage{algorithmic}
\usepackage{algorithm}
\usepackage{array}
\usepackage[caption=false,font=normalsize,labelfont=sf,textfont=sf]{subfig}
\usepackage{textcomp}
\usepackage{stfloats}
\usepackage{url}
\usepackage[dvipsnames]{xcolor}
\usepackage{hyperref}
\usepackage{verbatim}
\usepackage{graphicx}
\usepackage{cite}
\usepackage{multirow}
\usepackage[flushleft]{threeparttable}
\begin{document}

\title{Face Image Quality Enhancement Study for Face Recognition }
\author{Iqbal Nouyed, Na Zhang
\thanks{Iqbal Nouyed, Na Zhang are with Lane Department of Computer Science and Electrical Engineering at West Virginia University, Morgantown, WV 26506-6109. }
}

\maketitle

\begin{abstract}
Unconstrained face recognition is an active research area among computer vision and biometric researchers for many years now. Still the problem of face recognition in low quality photos has not been well-studied so far. In this paper, we explore the face recognition performance on low quality photos, and we try to improve the accuracy in dealing with low quality face images. We assemble a large database with low quality photos, and examine the performance of face recognition algorithms for three different quality sets. Using state-of-the-art facial image enhancement approaches, we explore the face recognition performance for the enhanced face images. To perform this without experimental bias, we have developed a new protocol for recognition with low quality face photos and validate the performance experimentally. Our designed  protocol for face recognition with low quality face images can be useful to other researchers. Moreover, experiment results show some of the challenging aspects of this problem.
\end{abstract}


\section{Introduction}
\label{sec:intro}

Although unconstrained face recognition has become an active research area among computer vision and biometric researchers in recent years, the problem of face recognition in low quality photos has not been well-studied yet. E.g. various face recognition approaches based on the LFW database \cite{Learned-Miller2016} does not consider the quality . Generally speaking, when we say a dataset has images taken in controlled environment, they are considered good quality (e.g., mugshot photos) images. However, when a dataset is constructed using images taken from unconstrained environment, it does not mean these images are all low quality. Figure~\ref{fig:lfw} below shows some examples of the LFW dataset with both high and low quality face images.

\begin{figure}[htb]
\includegraphics[width=\linewidth]{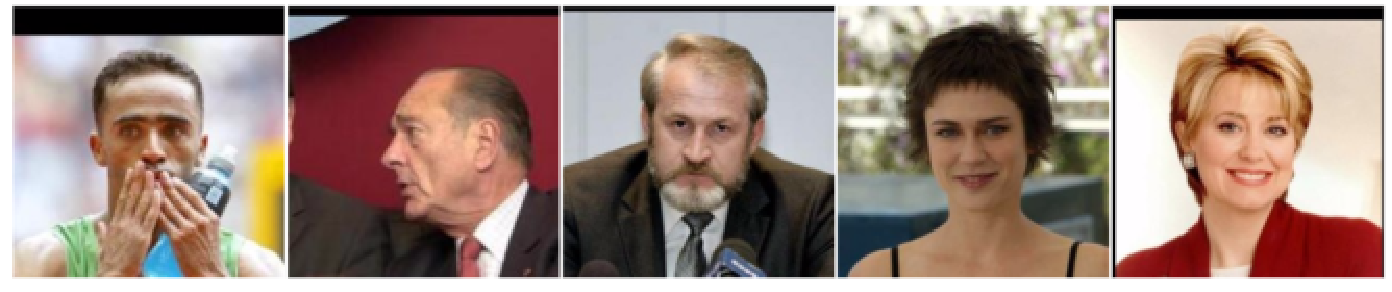}
\caption{Example of LFW images with different qualities.}
\label{fig:lfw}
\end{figure}

There have been many works published by researchers focusing on a single aspect of face image quality improvement. They have worked on improving resolution (e.g. super-resolution method) \cite{bilgazyev2011super,dahl2017super}, or pose correction (frontalization) \cite{huang2017rotation,ferrari2016frontalization,hassner2015frontal,sagonas2015frontalization}, or on blind image quality enhancement \cite{hait2017blind}. But so far we have searched there has never been a study on the combined affect of face image enhancement on biometric quality. Real world images can simultaneously have multiple quality attributes, e.g, having pose variation, low illumination and a large expression variation at the same image, which makes the problem very hard.

In this  paper, we focus on studying the affect of face image quality enhancement for improved face recognition with different image qualities. It is a very challenging problem and has not been well-studied yet. In this work, we have used a database of unconstrained face images and performed cross quality face recognition. Using our protocol we divided it into three different quality sets, namely, high, middle and low. Then, we tried to enhance the quality of the low and middle quality image sets by applying different image quality enhancement methods. The following sections describes the method in detail.


\section{Protocol}

\subsection{Database description}

The IARPA Janus Benchmark A (IJB-A) dataset \cite{klare2015ijba} is used for the study. it is a relatively new database with $21,230$ face images of $500$ subjects. It is a publicly available media in the wild dataset.  The IJB-A is a joint face detection and face recognition dataset. The dataset consists of face images and videos that were collected “in the wild”. A key distinction between this dataset and previous datasets is that all faces have been manually localized and have not been filtered by a commodity face detector. The implication of this approach is an unprecedented amount of variation in pose, occlusion and illumination in the IJB-A dataset.

\subsection{Face image quality assessment }

\begin{figure*}[htb]
\begin{center}
   \includegraphics[width=\linewidth,keepaspectratio]{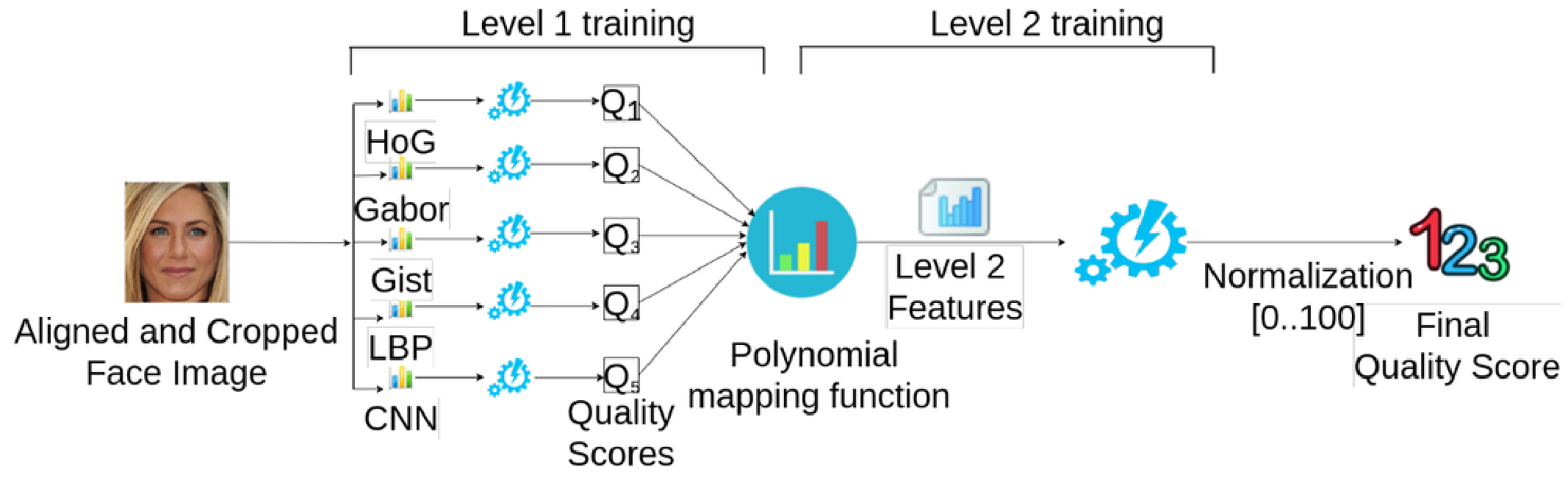}
\end{center}
   \caption{Two level learning method to calculate the face image quality.}
\label{fig:framework}   
\end{figure*}

A learning to rank based quality assessment approach proposed by Chen et al. \cite{chen2015fiqa} is used.  The face image quality framework uses two level training process to train a RankSVM. At level 1, we use five different face recognition features, namely, HoG, Gabor, Gist, LBP and CNN. At, level 2, we construct new features from the output of the first level prediction using a 5th degree polynomial kernel mapping function. The result of the second level prediction is normalized and rounded off and considered as the quality score.

Figure~\ref{fig:framework} shows the two-level learning process A dataset containing controlled, real world and non-face images is used for learning. The method is a good predictor of face image quality.

\subsection{Face recognition}

We use Openface face landmark algorithm \cite{baltru2016openface} to detect 68 face landmarks. Then, we use 6 face landmarks out of them (left and right eye corner, and mouth corners) to align the face image using the alignment and cropping method described in \cite{chen2015fiqa}. 



We extracted Gabor feature from the images. We considered high quality set as the gallery and, low and middle quality sets as probe sets. Used cosine similarity measure we matched the low and mid quality images with the high quality sets and calculated the recognition rate. 




\subsection{Quality filtering}

\begin{figure}[htb]
\includegraphics[width=\linewidth]{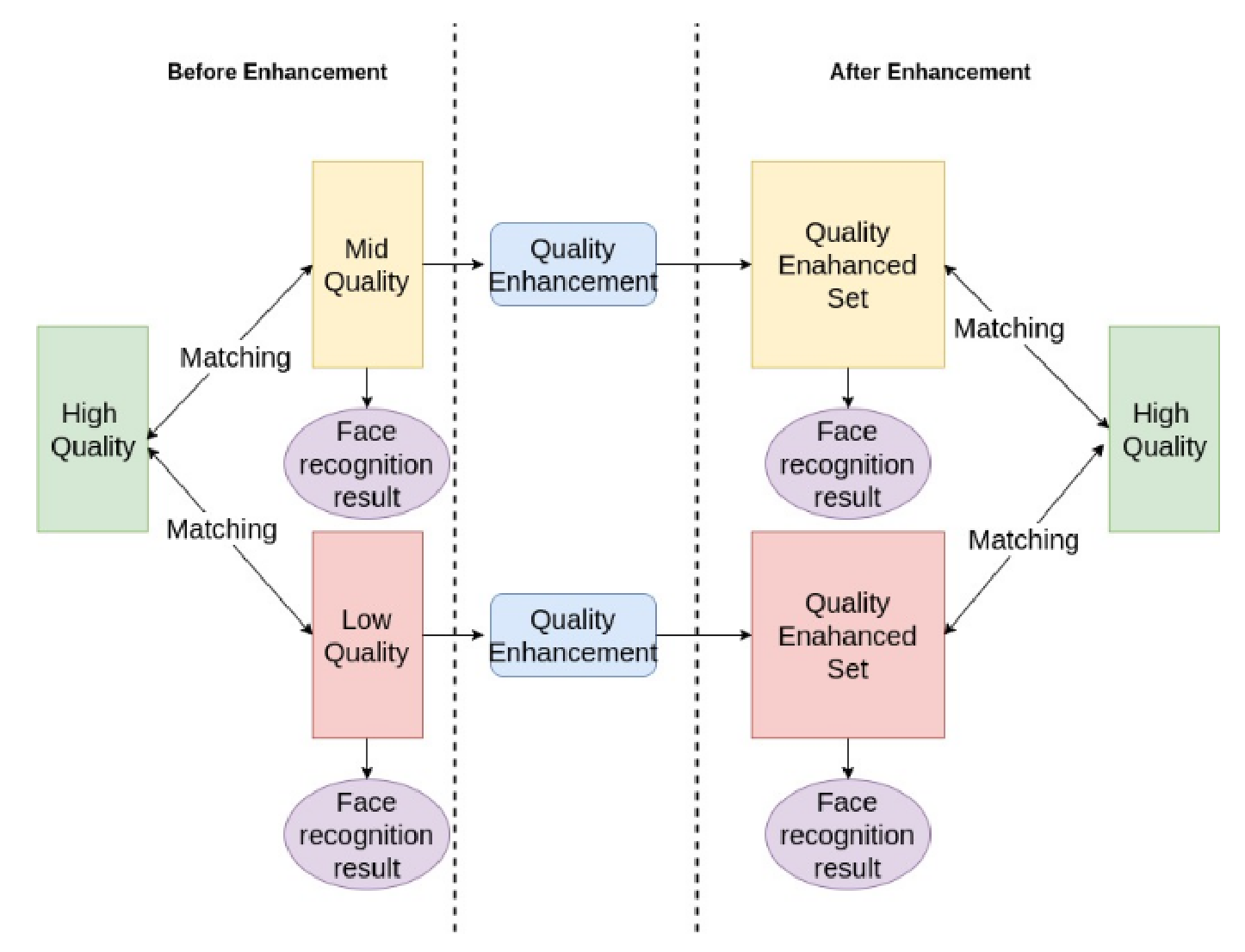}
\caption{Quality enhancement evaluation process overview.}
\label{fig:quality}
\end{figure}

 After cropping and alignment, we apply the face image quality assessment method on all the images and generate quality score for each of them. Then, we partition the database into three parts, based on our quality assessment. If $s$ is the quality score given for an image $I$ then the image is categorized as equation~(\ref{eq:category}) below.

\begin{eqnarray} \label{eq:category}
K =
\left\{
	\begin{array}{ll}
		\mbox{Low}  & \mbox{if } s < 30 \\
		\mbox{Middle} & \mbox{if } 30 \leq s < 60 \\
		\mbox{High} & \mbox{if } s \geq 60 \\		
	\end{array}
\right.
\end{eqnarray}

where $K$ is the face image category. 

Using high quality set as the gallery and the low and middle quality set as probe, we calculate the rank-1 to rank-50 face recognition accuracy for the low and middle quality sets. This we use as benchmark to compare with the face recogniton results of the enhanced low and mid quality sets later on.

Table~\ref{tab:database} shows the quality wise distribution for the IJB-A database. In, Figure~\ref{fig:quality} we show the overview of the quality enhancement evaluation process. 

\begin{table}[htb]
\begin{center}
\caption{Image distribution after quality filtering.} \label{tab:database}
\begin{tabular}{|l|c|c|c|c|}
  \hline
& High(H) &Middle(M) &  Low(L) \\
  \hline
 subjects & 500 & 483 &  489 \\
 images & 1,543 & 13,491 & 6,196 \\

  \hline

\end{tabular}
\end{center}
\end{table}


\section{Quality Enhancement}
There are various causes that can affect the quality of a face images, such as, pose variation, uneven or too high or too low illumination, image resolution, occlusion, motion blur etc. For our study we focused on three enhancement methods: 1) pose correction, 2) correcting motion blur and 3) normalizing illumination variation.

\subsection{Pose estimation and correction}

\begin{figure}[htb]
\includegraphics[width=\linewidth]{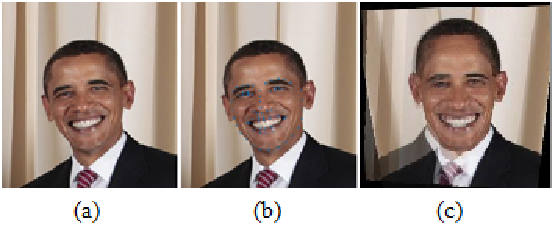}
\caption{Frontalization process overview. (a) Query photo; (b) facial feature detections; (c) frontalized result.}
\label{fig:frontalization}
\end{figure}

We use Openface \cite{baltru2016openface} to measure the head pose. OpenFace uses the Conditional Local Neural Fields (CLNF) model \cite{baltrusaitis2013constrained} which uses a 3D representation of facial landmarks and projects them to the image using orthographic camera projection. This allows accurate estimation of the head pose. Openface measure the head orientation in Euler angles. A head rotation can be characterized by three Euler angles: roll $(u)$, pitch $(v)$ and yaw $(w)$. Euler angles are commonly defined according to the right hand rule. Namely, they have positive values when they represent a rotation that appears clockwise when looking in the positive direction of the axis, and negative values when the rotation appears counter-clockwise. The angles are measured in radian, and thus has the range of the interval $[-\pi,\pi]$. \\

We chose the frontalization technique proposed by Hassner et al. \cite{hassner2015frontal} for pose correction. Banerjee et al. \cite{banerjee2016frontalize} have showed the frontalizing training and testing face images for deep features does not significantly improve the recognition performance. We want to investigate the affect of frontalization on face recognition using traditional features such as Gabor.\\

In this method, a face is first detected using an off-the-shelf face detector, and then cropped and rescaled to a standard coordinate system. Then facial feature points are localized and used to align the photo with a textured, 3D model of a generic, reference face. We use the Openface facial feature points detection method \cite{baltru2016openface} to detect the face landmarks. A rendered, frontal view of this face provides a reference coordinate system. The initial frontalized face is obtained by back-projecting the appearance of the query photo to the reference coordinate system using the 3D surface as a proxy. Then the final result is produced by borrowing appearances from corresponding symmetric sides of the face wherever facial features are poorly visible due to the query's pose. Figure~\ref{fig:frontalization} shows the overview of the frontalization process.

\subsection{Blur measure and deblurring}

\begin{figure}[htb]
\includegraphics[width=\linewidth]{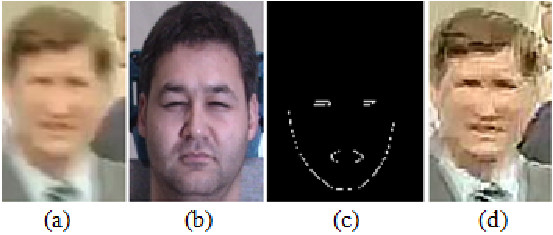}
\caption{Deblurring process overview. (a) Blurred face image; (b) exemplar image; (c) mask; (d) final result.}
\label{fig:deblur}
\end{figure}

To measure the blurriness of the face image we used two types of measures separately. First one was measureing the edge density. Edge density \cite{krotkov2012active} measures the average magnitude of the gradient over the face of a person. When images are in focus the average gradient magnitude will be higher than when the image is out of focus. This method has been shown to perform well at estimating the relative focus of images. As Beveridge et al. \cite{beveridge2008focus} we calculate edge density of the face region of the image only. For a given image, the edge density measures the average edge magnitude in a subregion of the image. Let $I(x,y)$ an image, and $e(x,y)$ is the edge magnitude of the image. For a subregion $r$ with the left-top corner at $(x_1,y_1)$ and the bottom-right corner at $(x_2,y_2)$ , the edge density measure $E$ is defined as:

\begin{eqnarray}
E = \frac{1}{a_r}\sum_{x=x_1}^{x_2}\sum_{y=y_1}^{y_2} e(x,y)
\end{eqnarray}

where $a_r$ is the region area, $a_r = (x_2-x_1+1)(y_2-y_1+1)$.

\begin{figure}[htb]
\includegraphics[width=\linewidth]{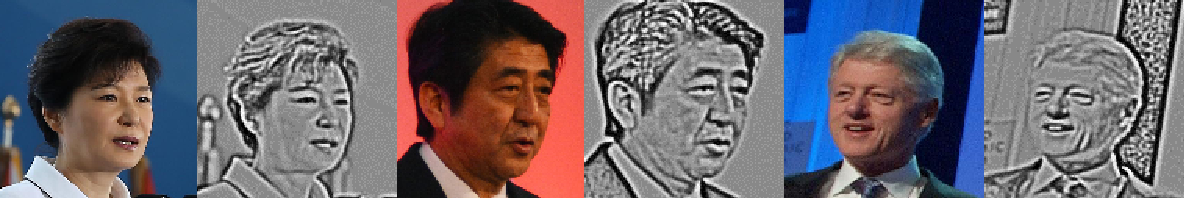}
\caption{Sample face images and their corresponding photometric normalization results.}
\label{fig:photo_norm}
\end{figure}

\begin{figure*}[htb]
\includegraphics[width=\linewidth]{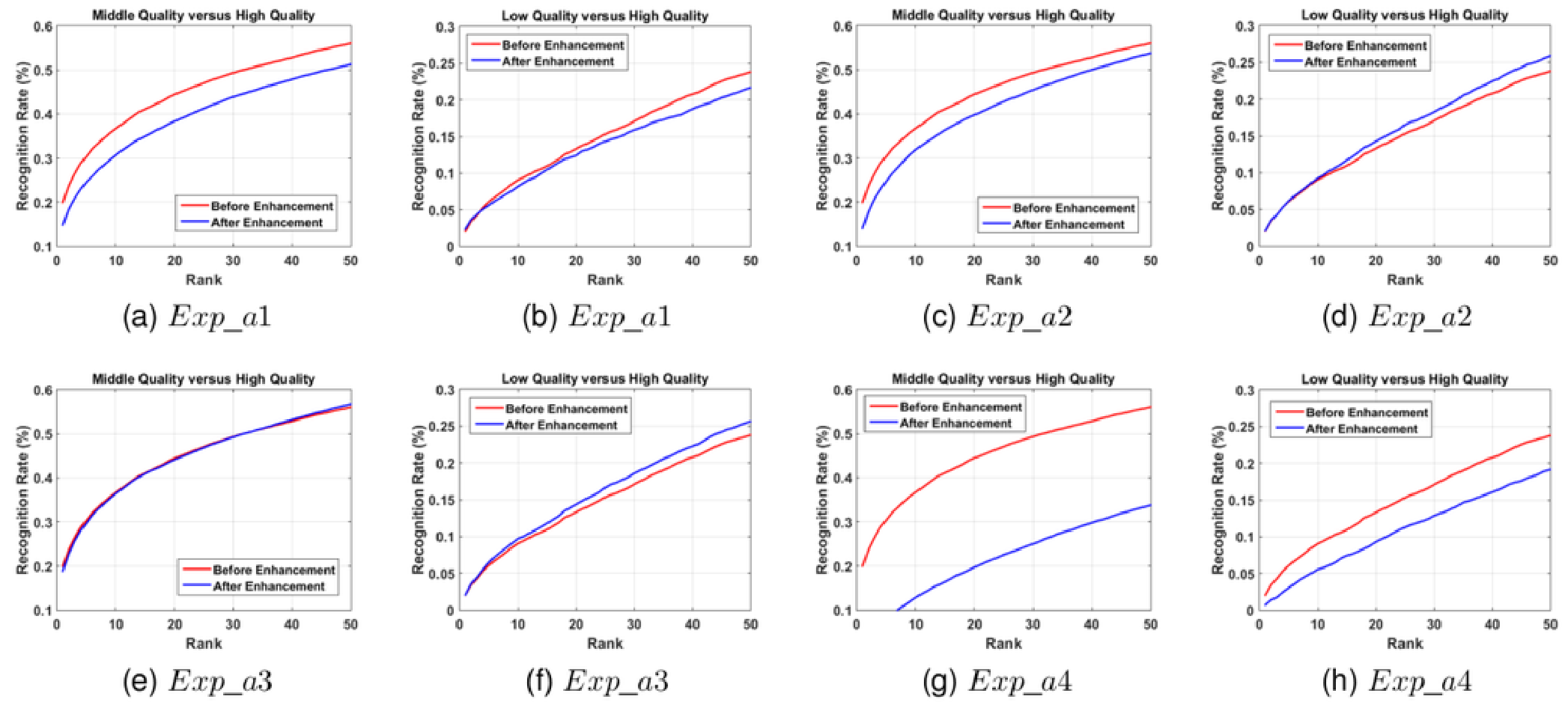}
\caption{CMC curves for enhancement by frontalization.}
\label{fig:cmc_frontal}
\end{figure*}

    

The second approach was to measure the sharpness. We applied a low-pass filter to the face image and then the average value of the pixels of the image is considered as the sharpness measure \cite{nasrollahi2008quality}.  It can be defined as: $\mbox{Sharpness} = \mbox{mean}(|I(x,y)-\mbox{Lowpass}(I(x,y))|) $, where $I(x,y) = $ is the intensity of the image at location $(x,y)$, $\mbox{Lowpass} = $ is lowpass Gaussian filter.




We chose an exemplar-based deblurring algorithm for face images proposed by Pan et al. \cite{pan2014deblurring} that exploits the structural information of the face. This method uses face structure and reliable edges from exemplars for kernel estimation. It is a maximum a posteriori (MAP) deblurring algorithm. We set, the kernel size, $k=13$, gamma correction parameter $g=1.0$, and the number of iterations, $i=50$ for the blind deconvolution step. Figure~\ref{fig:deblur} shows the deblurring process on a challenging example.

\subsection{Illumination measure and photometric normalization}

For some experiments we applied normalization on the entire image set or used illumination measure for thresholding the images. For measuring the illumination we use the spectral energy \cite{nill1992spectral}. It describes abrupt changes in illumination and specular reflection. The image is tessellated into several non-overlapping blocks and the spectral energy is computed for each block. The value is computed as the magnitude of Fourier transform components in both horizontal and vertical directions.

We did a study on 21 different photometric normalization methods for their effect on face recognition performance. We found that the method proposed by Wang et al. \cite{wang2011weberface} to be one of the top performing methods. This method uses the Weber's law, which concludes that stimuli are perceived not in absolute terms, but in relative terms. Given a face image , for each pixel we compute the ratio between two terms: one is the relative intensity difference of the current pixel against its neighbors; the other is the intensity of the current pixel. The obtained ratio is called ``Weber-face''. Weber-face can extract the local salient patterns very well from the input image, and it is an illumination insensitive representation.  The``Weber-face'' $W$ of an image $I(x,y)$ can be defined as:

\begin{equation}
\begin{split}
W(x,y) =& \mbox{arctan}\bigg(\alpha\sum_{i\in A}\sum_{j\in A}\frac{I(x,y)-I(x-i\Delta x,y-j\Delta y)}{I(x,y)}\bigg) 
\end{split}
\end{equation}

where $A=\{-1,0,1\}$, $\alpha$ is a parameter for adjusting the intensity difference betweeen neighboring pixels,  
We set, the standard deviation of the Gaussian filter used in the smoothing step, $\sigma=1$, the size of the neighborhood used for computing the Weberfaces, $n=9$, and, $\alpha=2$. Figure~\ref{fig:photo_norm} shows the normalization process on sample images.

\section{Experiment results}

 
%

\begin{figure*}[htb]
\includegraphics[width=\linewidth]{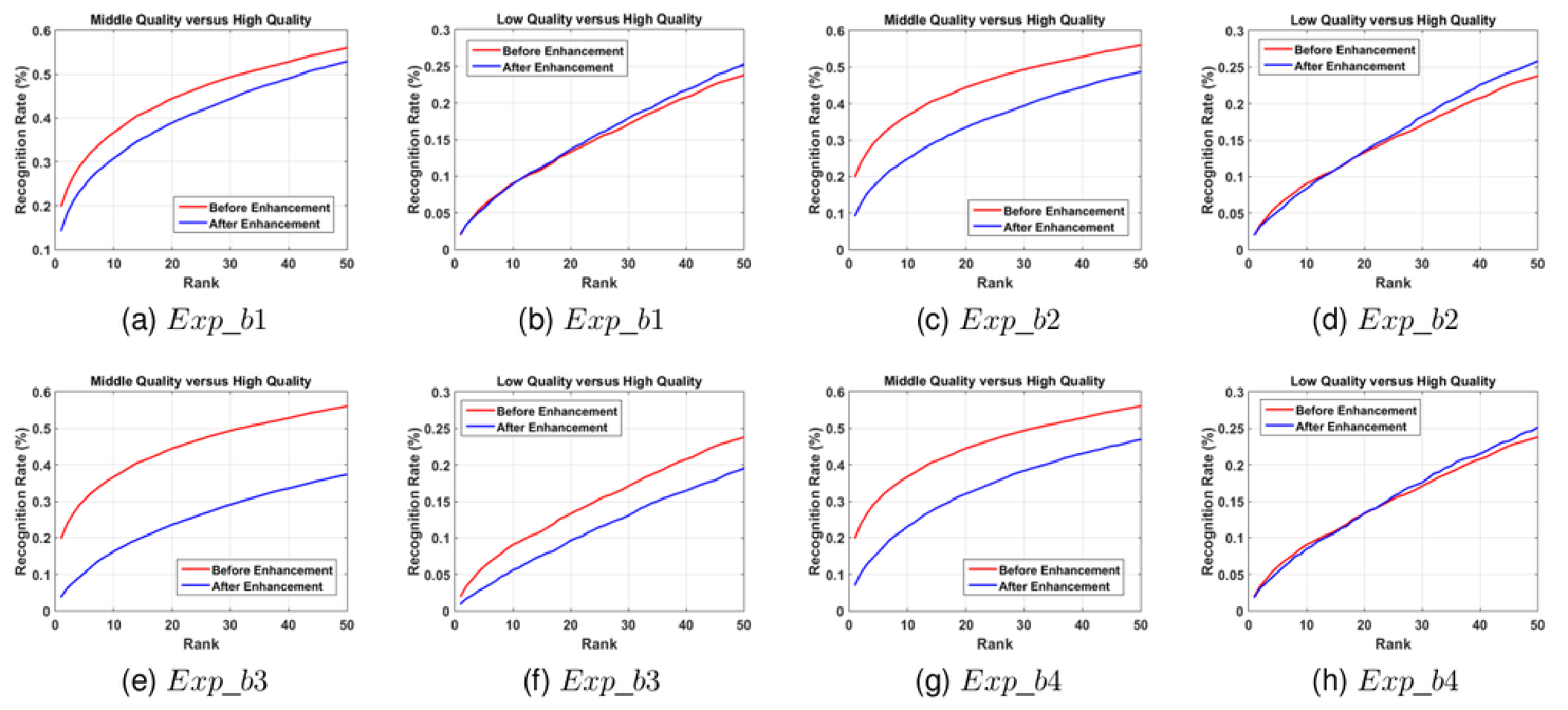}
\caption{CMC Curves for enhancement by deblurring.}
\label{fig:cmc_deblur}
\end{figure*}


\subsection{Pose correction}
We have performed several experiments to investigate the affect of pose correction on face image quality enhancement. For experiment $Exp\_a1$, we define the threshold of the three Euler angles (roll, pitch and yaw) as 30 degree. If one or more absolute values of three Euler angles of the face are greater than the threshold, this face will be selected to do frontalization.  We frontalized $1484$ images which is $11.00\%$ of the middle quality set and $3110$ images which is $50.19\%$ of the low quality set. We have found that when the pose is very large or the face is partially occluded the frontalization method failed to produce desired result, so we omit these faces. We adopt all faces in middle and low quality set to do face recognition including frontalized faces and the remaining faces. Figure~\ref{fig:cmc_frontal}(a) \& (b) show the result. We plotted the middle versus high quality and low versus high quality face recognition results for rank-1 to rank-50 to check if the enhancement affects the recognition rate. 


For $Exp\_a2$, we redefine the thresholds of the three Euler angles as the mean of absolute values of high quality image set which are $0.1263$ of pitch angle, $0.1432$ of yaw angle and $0.0947$ of roll angle. If one or more Euler angles of the face are greater than the thresholds, this face will be selected to do frontalization. We frontalized $11282$ images in middle quality set which is $83.63\%$ of the middle quality set and $4118$ images in low quality set which is $66.46\%$ of the low quality set. The frontalized faces are symmetric. We also adopt all faces including frontalized faces and the remaining faces to do face recognition. Figure~\ref{fig:cmc_frontal} (c) \& (d) shows the result.


In $Exp\_a3$, we redefined the thresholds of the three Euler angles. The threshold of roll is $45$ degree, the threshold of pitch is $25$ degree and the threshold of yaw is $15$ degree. If one or more absolute values of Euler angles of the face are greater than the thresholds, then the face is selected for frontalization. We use symmetric frontalized faces for face recognition. We frontalized $3783$ images in middle quality set which is $28.04\%$ of the middle quality set and $1907$ images in low quality set which is $30.78\%$ of the low quality set. Figure~\ref{fig:cmc_frontal} (e) \& (f) shows the result. 


Similar to $Exp\_a3$, experiment $Exp\_a4$ uses the same thresholds to perform frontalization, but it adopts all images. That means, including both frontalized faces and the remaining faces to perform face recognition. Figure~\ref{fig:cmc_frontal}(g) \& (h) shows the result. 


Among all the results, experiment $Exp\_a3$ (Figure 9) produces the best performance. Both middle and low versus high quality experiments have improvements in image recognition rate and the improvement for low versus high quality set is more significant. In low versus high quality experiment of $Exp\_a2$ (Figure 8), the recognition rate also has big improvement. For $Exp\_a1$ and $Exp\_a4$, the recognition rates of both middle and low versus high quality set are not increased. 

\subsection{Deblurring}


\begin{figure*}[htb]
\includegraphics[width=\linewidth]{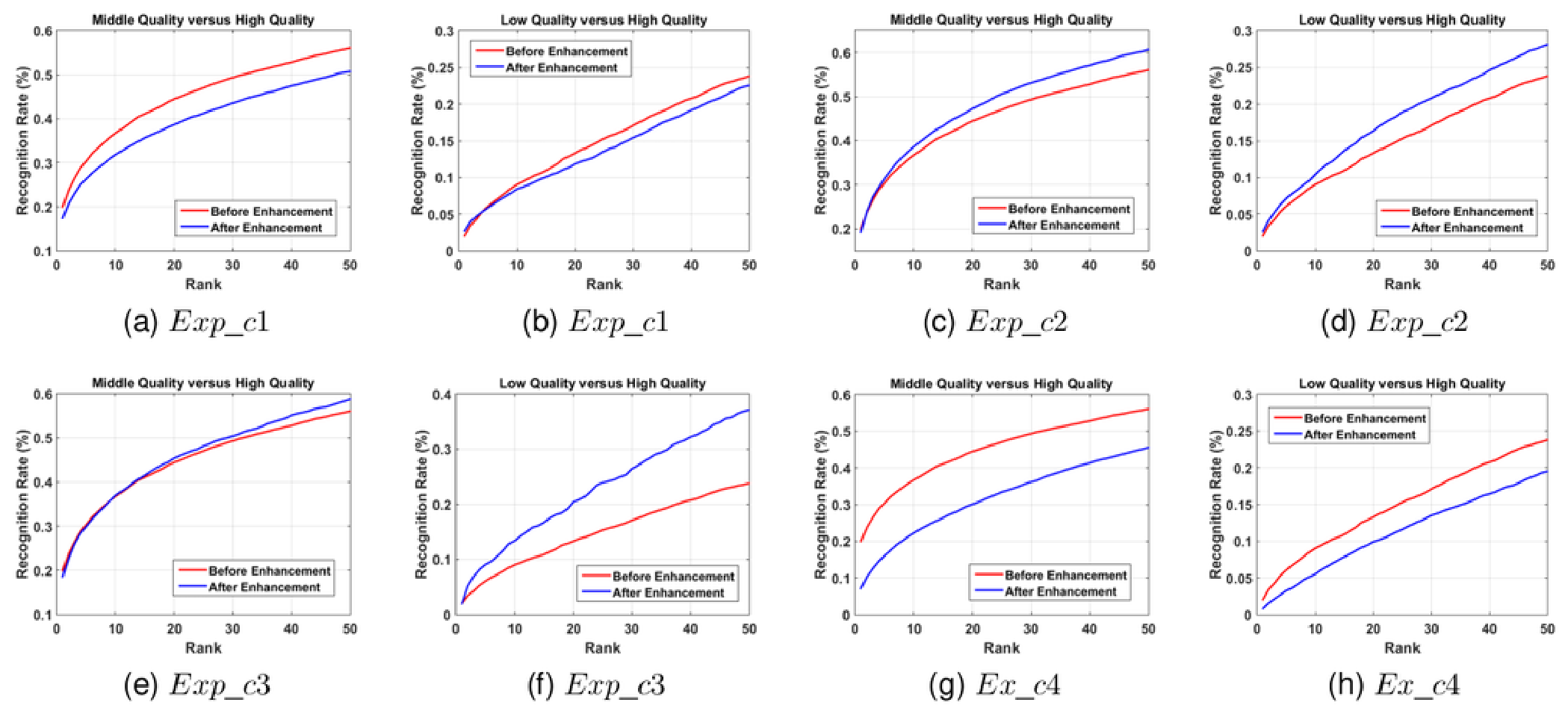}
\caption{CMC curves for enhancement by photometric normalization.}
\label{fig:cmc_photo}
\end{figure*}

Several experiments have also been performed to investigate the affect of deblurring on face quality enhancement. Before deblurring, we need the images of frontalization evaluation process (pose correction experiment $Exp\_a3$) since we adopt an exemplar based deblurring method \cite{pan2014deblurring} for deblurring. This method only has frontal faces in their exemplar image set. So, pose correction is needed before we can match the images to find the appropriate exemplar for the input image. We use the cropped and aligned images of the frontalization evaluation process and remaining images in middle and low quality set as the input of the deblurring process.  In $Exp\_b1$, we using mean focus measure of the high quality set as the threshold ($30.802$). Using this we divide the input set into two parts, one to be deblurred another we keep intact. After deblurring, we extract the features of all images including deblurred and frontalized images (set 1), deblurred but not frontalized images (set 2) and remaining images (set 3) in middle and low quality set as shown in Table~ \ref{tab:table_exp_b1_b3}. And then face recognition is performed. 

\begin{table}[htb]
\begin{center}
\caption{Number of images for each quality sets in $Exp\_b1$ and $Exp\_b3$.} \label{tab:table_exp_b1_b3}
\begin{tabular}{|l|l|c|c|c|c|}
  \hline
Exp. & Quality & \#set1 & \#set2 & \#set3 & In total \\
  \hline
 $Exp\_b1$ & mid & 6328 & 1952 & 5211 & 13491 \\
  & low & 2032 & 2564 & 1600 & 6196 \\
  \hline
$Exp\_b3$ & mid & 2166 & 4943 & 6382 & 13491 \\
 & low & 780 & 2006 & 3410 & 6196 \\
  \hline	  
\end{tabular}
\end{center}
\end{table}

We plotted the middle versus high quality and low versus high quality face recognition results for rank-1 to rank-50 to check if the enhancement process affects on the recognition rate. Figure~\ref{fig:cmc_deblur} (a) \& (b) show the results.


For $Exp\_b2$, we use all faces as our input including frontalized faces and the remaining, and use focus measure threshold to get image list of deblurring. But we, after deblurring, extract the features of equal number (n=461) of subjects of middle and low deblurred images which are $8280$ images of middle quality and $4359$ images of low quality.  And then face recognition is performed. Figure~\ref{fig:cmc_deblur} (c) \& (d) show the results.


Similar to $Exp\_b1$, in $Exp\_b3$ we just change the threshold to the mean value of sharpness ($0.15444$) of high quality images. After deblurring, we also extract the features of all images including deblurred and frontalized images (set1), deblurred but not frontalized images (set2) and remaining images (set3) in middle and low quality set as shown in Table~\ref{tab:table_exp_b1_b3}. And then face recognition is performed. Figure~\ref{fig:cmc_deblur} (e) \& (f) show the results.


Similar to $Exp\_b3$,  in experiment $Exp\_b4$, we also use mean value of sharpness ($0.15444$) of high quality images as our threshold to get image list of deblurring. After deblurring, however, we just extract the features of deblurred images. And then face recognition is performed. Figure~\ref{fig:cmc_deblur} shows the results.


We can see that for low versus high quality set of $Exp\_b1$, $Exp\_b2$ and $Exp\_b4$, there is an improvement in recognition rate which indicates the deblurring process increases the quality of the low images. 


\begin{figure*}[htb]
\includegraphics[width=\linewidth]{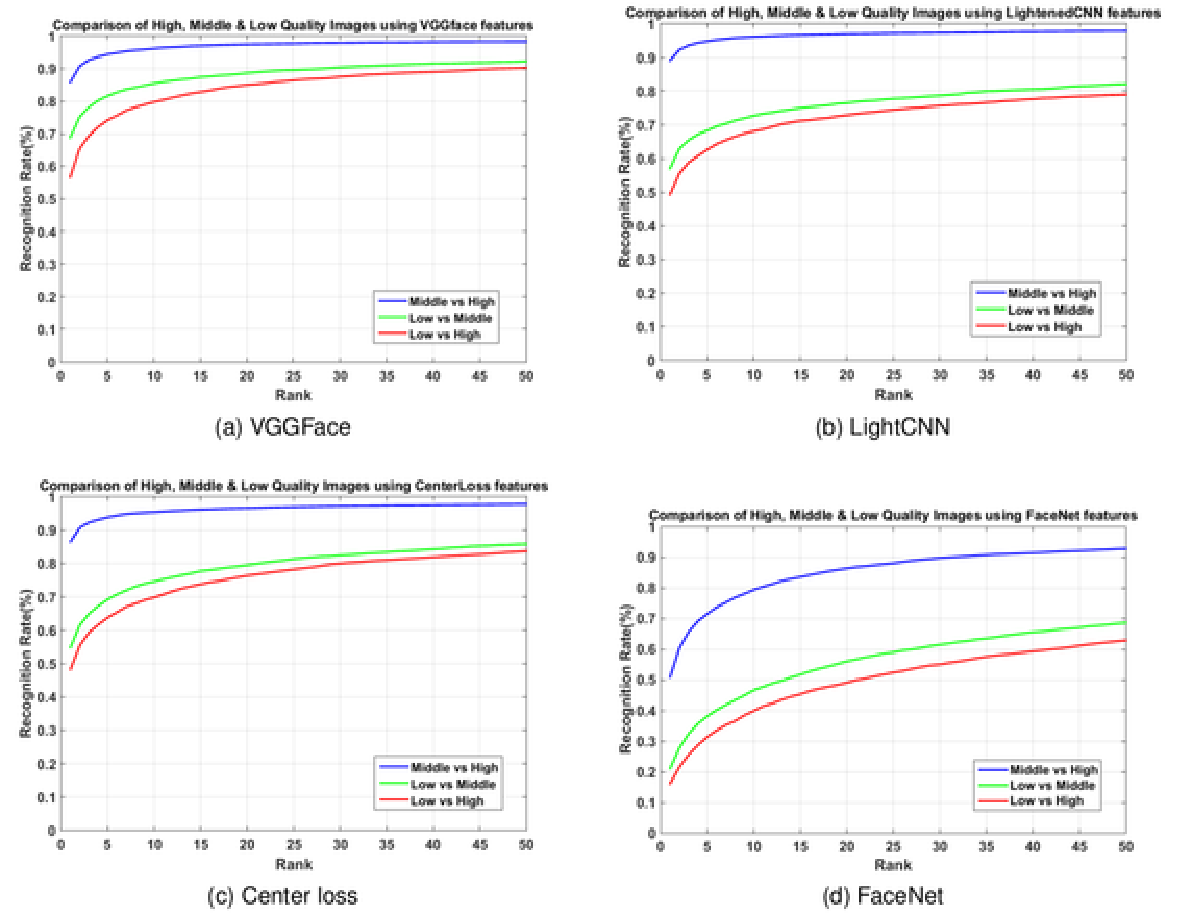}
\caption{CMC curves for face recognition performance for different deep features.}
\label{fig:cmc_diff_deep}
\end{figure*}

\subsection{Photometric normalization}
Several experiments have also been performed to investigate the affect of photometric normalization on face quality enhancement. In experiment $Exp\_c1$, we take all original images of the high, middle and low quality image sets as input of photometric normalization process. Then we use photometric normalized faces to perform landmark detection, cropping and alignment and then extract feature to do face recognition. We plotted the middle versus high quality and low versus high quality face recognition results for rank-1 to rank-50 to check if the enhancement affects the recognition rate. Figure~\ref{fig:cmc_photo} (a) \& (b) show the results. 


Similar to experiment $Exp\_c1$, in experiment $Exp\_c2$ we also take all images of high, middle and low quality image set as input of photometric normalization process. The difference is that we use cropped faces of original images in this experiment. Figure~\ref{fig:cmc_photo} (c) \& (d) show the result. 


The evaluation process of experiment $Exp\_c3$ takes the original images of the high, middle and low quality sets as input. We apply the chosen illumination normalization method on middle and low quality images.  The mean value of illumination of the faces in high quality image set is used as the threshold ($59503$). In middle and low quality image sets, if the value of illumination measure of the face is greater than the threshold, this face will be selected to do photometric normalization. Then we use photometric normalized faces ($7550$ images in middle quality set and $3326$ images in low quality set) to perform landmark detection, cropping and alignment. As before, we extract feature and find recognition rates.  
We found that, for some cases when the illumination is very high near or inside the face recognition, it obliterates the face information and the normalization process cannot improve the quality. Figure~\ref{fig:cmc_photo} (e) \& (f) shows the results. 


Similar procedure with experiment $Exp\_c3$, experiment $Ex\_c4$ use same threshold value of illumination measure. The difference is that $Exp\_c4$ use both photometric normalized and the remaining images of middle and low quality image sets to do face recognition. Figure~\ref{fig:cmc_photo} (g) \& (h) shows the results.


For experiment $Exp\_c3$ and $Exp\_c2$, both middle and low versus high quality image set experiments have significant improvement especially low versus high quality. For experiment $Exp\_c1$ and $Exp\_c4$, the recognition rates of both middle and low versus high quality set are not increased.

\subsection{Enhancement by deep feature}


\begin{figure*}[htb]
\includegraphics[width=\linewidth]{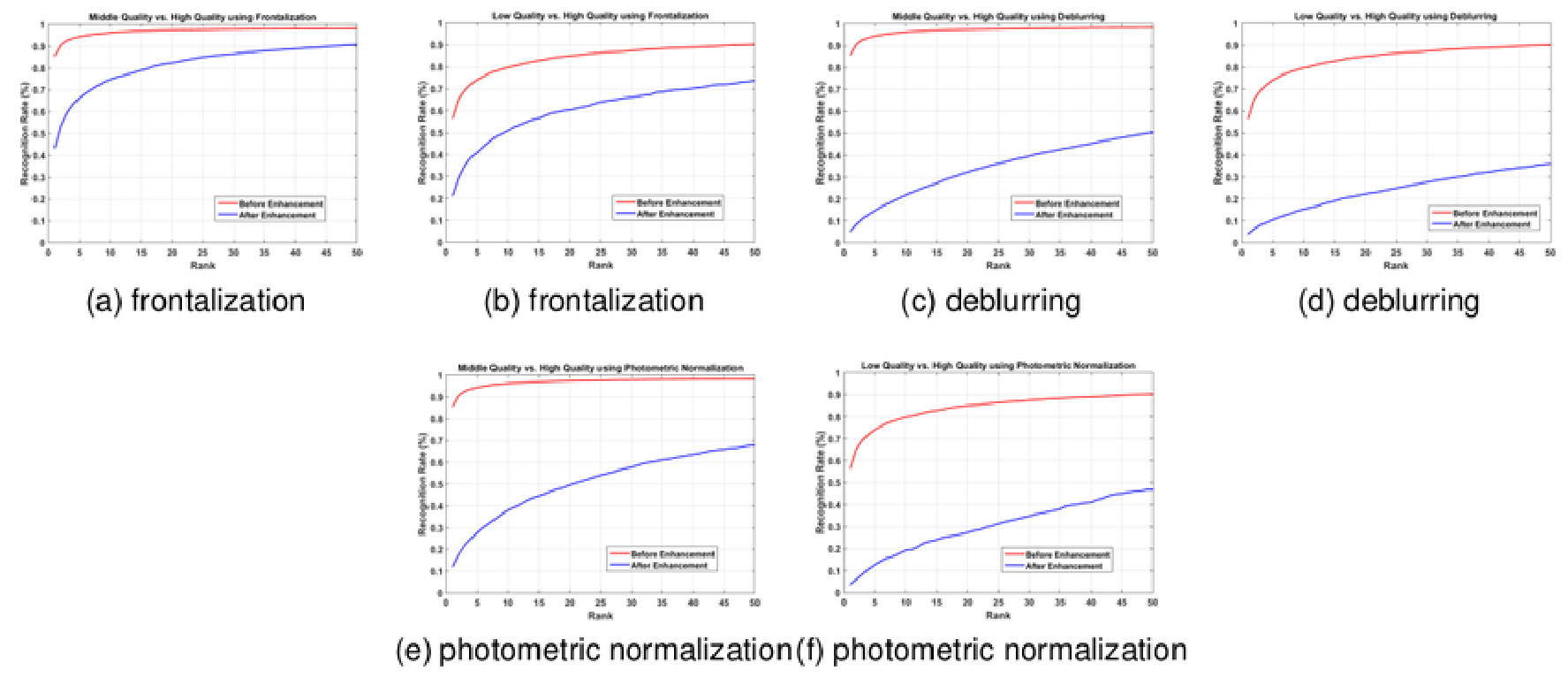}
\caption{CMC curves for deep feature based face image quality enhancement experiments.}
\label{fig:cmc_deep}
\end{figure*}

Four different pre-trained deep models were used to study their performance on different quality sets: VGGFace \cite{parkhi2015deep}, LightCNN\cite{wu2015lightened}, Center loss \cite{wen2016discriminative}, and FaceNet\cite{schroff2015facenet}. In Figure~\ref{fig:cmc_diff_deep} we can see their performance plots. It can be observed that the VGGFace has the overall good performance among different qualities, thereby it was chosen for the enhancement experiment. 

We choose the processed images with best performance ($Exp\_a3$, $Exp\_b1$ and $Exp\_c3$) among all experiments for each method and extract feature using VGGFace~\cite{parkhi2015deep} deep model. This could be because of VGGFace having the largest feature size of $4096$ containing more face quality information comparative to other deep features. Figure~\ref{fig:cmc_deep} (a) \& (b) shows the result of pose correction, Figure ~\ref{fig:cmc_deep} (c) \& (d) shows the result of deblurring and Figure ~\ref{fig:cmc_deep} (e) \& (f) shows the result of photometric normalization. We can find that the recognition rate of after enhancement is much lower than that of before enhancement. That the processed images are morphed more than original images for deep model may be the major reason.




\subsection{Combination}
We have also tried to combine three methods (pose correction, deblurring and photomrtric normalization) to enhance faces. We find that for these three methods, different measures, different values of threshold and the number of images used to do face recognition affect the recognition result a lot. So, there is a lot of parameter tuning involved which makes it very complicated to combine these three methods, but we intend to investigate on more about this in the future.


\section{Conclusion}
As the recognition rates in the plot figures indicate, it is extremely challenging to improve face recognition performance of low quality photos. In this work we have shown, that some enhancement techniques can improve the face recognition accuracies, while some others may not. In our current evaluation, frontalization, deblurring and photometric normalization techniques can enhance the face image quality, though the improvement is not significant ye for all cases.  We need to investigate more, on how to find the right enhancement methods, and improve on the existing ones in future work to improve these results. Finally, much more efforts are needed to understand how to improve the quality of the face images to gain high accuracy in low-quality face recognition.

\bibliographystyle{IEEEtran}
\bibliography{face_enhancement}

\end{document}